\documentclass[sigconf]{aamas}

\usepackage{amsmath,amsfonts,bm}

\def\eqref#1{equation~\ref{#1}}

\def\1{\bm{1}}

\def\rx{{\textnormal{x}}}

\DeclareMathAlphabet{\mathsfit}{\encodingdefault}{\sfdefault}{m}{sl}
\SetMathAlphabet{\mathsfit}{bold}{\encodingdefault}{\sfdefault}{bx}{n}

\newcommand{\E}{\mathbb{E}}

\usepackage{balance} 
\usepackage{subcaption}
\usepackage{gensymb}
\usepackage{subcaption}
\usepackage{graphicx}
\usepackage{microtype}
\usepackage{placeins}
\usepackage{hyperref}
\usepackage{algorithm}
\usepackage[noend]{algpseudocode}

\setcopyright{ifaamas}
\acmConference[AAMAS '21]{Proc.\@ of the 20th International Conference on Autonomous Agents and Multiagent Systems (AAMAS 2021)}{May 3--7, 2021}{London, UK}{U.~Endriss, A.~Now\'{e}, F.~Dignum, A.~Lomuscio (eds.)}
\copyrightyear{2021}
\acmYear{2021}
\acmDOI{}
\acmPrice{}
\acmISBN{}

\acmSubmissionID{555}

\title[AAMAS-2021 Formatting Instructions]{Emergent Communication under Competition}

\author{Michael Noukhovitch}
\affiliation{
  \institution{Mila, Universit\'e de Montr\'eal}
  \city{Montr\'eal, Canada}}
\email{noukhovm@mila.quebec}

\author{Travis LaCroix}
\affiliation{
  \institution{Mila, Universit\'e de Montr\'eal \\ Schwartz Reisman Institute, University of Toronto}
  \city{Montr\'eal, Canada and Toronto, Canada}}
\email{lacroixt@mila.quebec}

\author{Angeliki Lazaridou}
\affiliation{
  \institution{Deepmind}
  \city{London, UK}}
\email{angeliki@google.com}

\author{Aaron Courville}
\affiliation{
  \institution{Mila, Universit\'e de Montr\'eal \\ CIFAR Fellow}
  \city{Montr\'eal, Canada}}
\email{aaron.courville@umontreal.ca}

\begin{abstract}
The literature in modern machine learning has only negative results for learning to communicate between competitive agents using standard RL. We introduce a modified sender-receiver game to study the spectrum of partially-competitive scenarios and show communication can indeed emerge in a competitive setting. We empirically demonstrate three key takeaways for future research. First, we show that communication is proportional to cooperation, and it can occur for partially competitive scenarios using standard learning algorithms. Second, we highlight the difference between communication and manipulation and extend previous metrics of communication to the competitive case. Third, we investigate the negotiation game where previous work failed to learn communication between independent agents \citep{Cao-et-al-2018}. We show that, in this setting, both agents must benefit from communication for it to emerge; and, with a slight modification to the game, we demonstrate successful communication between competitive agents. We hope this work overturns misconceptions and inspires more research in competitive emergent communication. Code is available at \url{http://github.com/mnoukhov/emergent-compete}
\end{abstract}


\newcommand{\BibTeX}{\rm B\kern-.05em{\sc i\kern-.025em b}\kern-.08em\TeX}

\begin{document}


\pagestyle{fancy}
\fancyhead{}


\maketitle


\section{Introduction}

Understanding the principles involved in the evolution of effective human communication is essential for the progression of artificial-intelligence research since following these same principles is likely our best path forward in developing effective communication strategies for use by interacting AI agents. Naturally, AI agents need a common language to coordinate with one another and to communicate successfully with humans \citep{Wagner-2003}. The emergence of communication protocols between learning agents has seen a surge of interest in recent years, but most work to date focuses on fully-cooperative agents that share rewards \citep{Foerster-et-al-2016,Havrylov-Titov-2017,Lazaridou-et-al-2016}. Work on partial or total conflict of interest between agents has been limited, with only negative results in modern ML literature \citep{lazaridou2020-communication82}. Results suggest that when agents are not fully cooperative, they will fail to learn to use a communication channel effectively \citep{Cao-et-al-2018} without extra explicit incentives to do so \citep{Jaques-et-al-2018}. This has contributed to a potential misconception that emergent communication is a purely cooperative pursuit \citep{lanctot2017}. Contrary to this view, theoretical results in classical game theory prove it is possible to use an existing cheap-talk communication channel to effectively coordinate actions even in situations where there is (partial) conflict of interest \citep{Farrell-Rabin-1996}.

In this paper, we aim to reconcile these contradictory views and lay the groundwork for future research in competitive emergent communication. We begin by reviewing the field of emergent communication and previous work in competitive cases.
Then, to study this problem in detail, we consider the simplest case of communication, where a sender must communicate a target to a receiver using a message. The receiver uses the message to guess the sender's target \citep{Lewis-1969}.
The classic game is fully cooperative, with agents sharing reward proportional to the accuracy of the guess. To investigate whether meaningful communication can emerge between {\it competitive} agents, we introduce degrees of conflict between the sender and receiver using a parameter that smoothly controls the distance between agents' targets. This allows us to study emergent communication
along a continuum between fully-competitive and fully-cooperative agents.

We present three key results for the field of emergent communication and provide takeaways for each.
\begin{enumerate}
    \item For the first time in modern emergent communication, we find evidence that agents can learn to effectively communicate in competitive scenarios---given that their interests are at least partially aligned.
    \item We investigate what qualifies as `communication' and distinguish it from `manipulation', leaning on previous literature in evolutionary game theory. In doing so, we extend the work of measuring cooperative communication \citep{Lowe2019OnTP} to the competitive case.
    \item We go back to one of the first instances of competitive emergent communication in \citet{Cao-et-al-2018} to reproduce and analyze their results. We bring forward a strong argument to overturn their hypothesis that agents fail to learn to communicate. Instead, we propose that agents specifically learned to communicate ineffectively, and we highlight that both agents must be able to benefit from communication to ensure its feasibility.
\end{enumerate}
Finally, we apply these three lessons to the negotiation game \citep{Cao-et-al-2018} where competitive agents previously failed to communicate. With a small modification to the game, we show that competitive agents learn to effectively communicate. We conclude with a view towards the future of the field and encourage future research to use the outlined takeaways to carefully build better models, environments, and metrics.

\section{Related Work}

\subsection{Emergent Communication}
Emergent communication (EC) is the study of learning a communication protocol, typically in order to share information that is needed to solve some task \citep{Foerster-et-al-2016}. Recent work has focused on deep learning \citep{Goodfellow-et-al-2016} agents trained with reinforcement learning (RL) \citep{Sutton1998}. Investigations have remaining mostly in the realm of fully-cooperative games \citep{Lazaridou-et-al-2018, Das2019TarMac, Evtimova-et-al-2017} with agents often communicating through discrete tokens from a vocabulary \citep{Havrylov-Titov-2017}. See \citet{lazaridou2020-communication82} for a full review of recent work.

Evaluating communicative success is often qualitative \citep{Lazaridou-et-al-2016, Havrylov-Titov-2017} and varied \citep{Bogin-et-al-2018}. \citet{Lowe2019OnTP} investigate common issues in evaluation and find that improvement over a non-communication baseline is sufficient to demonstrate success in \textit{non-situated} environments \citep{Wagner-2003} where agents can only transfer information to each other through communication actions. They also propose a metric for situated games, Causal Influence of Communication (CIC), which measures the causal influence of one agent's communication on another agent's action. A similar metric is used as an auxiliary reward by \citet{Jaques-et-al-2018} and \citet{eccles2019-communication9} to improve performance in multiagent RL games. In both cases there is an underlying assumption that more influence means better communication; however, this assumption may not always hold in competitive environments.

\subsection{Competitive Emergent Communication}
We investigate the space of partially-competitive games, known as general-sum games, where there is some amount of cooperation and some amount of competition. Research on EC in these environments has been limited with many works using forms of competition but not actually having game-theoretic competitiveness. \citet{singh2018learning} claimed to learn communication using continuous signals (scalars) in mixed cooperative-competitive scenarios. However, their setup uses parameter sharing between opponents; their `mixed' case is non-competitive (and implicitly cooperative); their competitive case is actually two stages---one fully cooperative and one fully competitive; and their result in the competitive scenario is essentially non-communication.\citet{liang2020-communication27} have two teams `competing' to solve a task but in practice they are actually cooperative, helping each other bootstrap their communication, and not competing over the same reward.

There are two modern works that do look at EC in competitive games.
\citet{Cao-et-al-2018} primarily focus on fully-cooperative agents that share a reward (`prosocial') but have a set of experiments with independent agents that do not share a reward (`selfish'). They find that independent agents playing a non-iterated, two-player negotiation game fail to learn to communicate effectively. They claim that agents that share a reward learn to communicate, whereas agents that do not may require an iterated game to learn effective communication. \citet{Jaques-et-al-2018} focused on many independent agents playing a multi-agent sequential social dilemmas (SSDs) \citep{Leibo-et-al-2017} and propose an auxiliary influence reward to encourage communication and cooperation. They show that effective communication between competitive agents can be learned using their auxiliary reward, but results without it are weak.

We focus on the rigorous study of why and when communication is learned in competitive settings. In contrast to previous work, we seek not only to learn communication between competitive agents but also to explain sufficient conditions for it while accounting for many possible variables and confounding factors. Our work uses a two-player non-iterated game as in \citet{Cao-et-al-2018}; but, we delve further and evaluate the spectrum from fully cooperative to fully competitive. To do so, we use independent agents (`selfish' in \citet{Cao-et-al-2018} terminology) and a game environment where we can precisely control the level of competition. Compared to \citet{Cao-et-al-2018}, we specify the level of competition, and probe agent behaviours as well as game dynamics to understand the underlying incentive structure. In contrast to \citet{Jaques-et-al-2018}, we use a non-situated environment to guarantee no communication through the action space, we explicitly quantify our game's level of competitiveness, and use a foolproof non-communication baseline. This work does not propose new architectures or learning rules but aims to critically examine existing beliefs, explain previous negative results, draw important distinctions, and recommend best practices for future work.

We follow previous work by comparing \textit{learning algorithms} that are trained together as opposed to \textit{learned agents} compared at test time. This is to avoid a significant issue of comparing EC agents---i.e., different protocols. Since an emergent protocol only has meaning between the agents that learned it together, comparing two learned agents at test time would require that they ad-hoc infer each others' protocols. While ad-hoc play has made strides in fully cooperative environments \citep{hu2020otherplay}, it has not been generally solved \citep{Bard2019TheHC}.

\subsection{Sender-Receiver Games}
Our experiments use the simple emergent-communication framework known as a sender-receiver game \citep{Lewis-1969} (or `referential game' \citep{Lazaridou-et-al-2016}).
In the classic game, the sender is given a target value to be communicated to the receiver via a message. The receiver receives the sender's message and must decode it to predict the target value. Both players get the same reward: the negative of the receiver's prediction error.
In this fully-cooperative setting, players learn a protocol to transfer information as effectively as possible \citep{Skyrms-2010-Signals}.
The messages between the sender and the receiver can be categorised as `cheap talk': messages are costless, non-binding, non-verifiable and \textit{may} affect the receiver's beliefs \citep{Farrell-Rabin-1996}.

Our work benefits greatly from \citet{Crawford-Sobel-1982}---a seminal work in classical game theory. They study possible \textit{fixed} communication equilibria under competition by giving the sender and receiver different targets and creating a conflict of interest. They perform a static analysis and prove the existence of Nash equilibria where the amount of information communicated is proportional to the alignment between the players' interests; however, no informative equilibrium exists when interests diverge too significantly. %
However, static analysis only tells us which equilibria are possible, not necessarily which are feasible to achieve and maintain using learning algorithms (e.g. RL). \citet{LaCroix-2020} provides a concrete case where static analysis fails to explain the properties of the game fully.  Behaviour out of equilibrium can sometimes be surprising; for example, \citet{Wagner-2012} proves that non-convergent dynamics can sustain partial information transfer even in a zero-sum signalling game. %

As such, we do not characterise the static equilibria of our game; instead, we provide  a {\it dynamic} analysis and show the \textit{feasibility} of communication using standard learning rules in RL. We do not explicitly aim for equilibria \citep{Shoham2003MultiAgentRL} but look at the information transfer of communication protocols in flux (and therefore out of equilibrium). This is more in line with previous work in EC \citep{Cao-et-al-2018, Jaques-et-al-2018} and allows us to give practical advice for training RL agents to communicate. Similarly, our work is related to the competitive sender-receiver games of \citet{kamenica2011bayesian} who make the assumption of rational, Bayesian agents whereas we work in field of modern EC which uses RL and deep learning \citep{lazaridou2020-communication82}.

\begin{figure}
    \centering
    \includegraphics[width=0.6\linewidth]{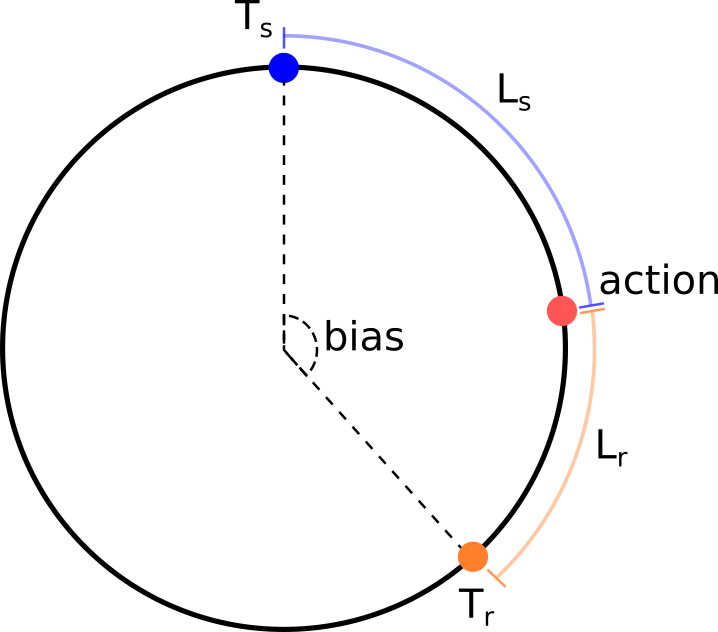}
    \caption{Visual representation of the circular biased sender-receiver game. Agents are given targets $T_R, T_S$ that are bias $b$ apart. The receiver chooses an action $a$ and agents receive the $L_1$ losses $L_1^r, L_1^S$}
    \label{fig:sender-receiver}
\end{figure}

\section{The Circular, Biased Sender-Receiver Game}

To investigate a range of competitive scenarios, we introduce a modified sender-receiver game with a continuous bias variable, $b$, that represents the agents' conflict of interest, ranging from fully cooperative to fully competitive. The two players---the Sender ({\it S}) and the Receiver ({\it R})---have corresponding targets ($T_s$ and $T_r$), which are represented by angles on a circle that are $b$ degrees apart: $T_r = (T_s + b) \mod 360\degree$.

The game starts with the sender's target being sampled uniformly from the circle $T_s \sim \mathrm{Uniform}[0,360)$. The sender is given its target as input and outputs a categorical distribution $S(T_s)$ over the vocabulary $V$. For its message, it stochastically samples a single token from the distribution $m \sim S(T_s)$. The receiver is given the message and deterministically outputs a scalar $a = R(m)$ as its action. The receiver does not see its own target so must rely on information from the sender's message. The goal of each agent is to make the receiver's action as close as possible to its {\it own} target value as both players get a loss proportional to the distance between the action and their respective targets. We choose to use an $L_1$ loss which corresponds to the angle between the target and action so formally, the loss for each agent i is

\begin{equation*}
    L_1^i(a, T_i) = \min(|T_i-a|, 360\degree-|T_i-a|)
\end{equation*}

Figure~\ref{fig:sender-receiver} gives an instance of this game; see Algorithm \ref{alg:game} in supplementary for the complete algorithm. It is evident that a game with $b = 0\degree$ is fully cooperative ($L_1^r = L_1^s$) and a game with the maximum bias $b = 180\degree$ is  fully competitive or constant-sum, a generalisation of zero-sum (see supplementary~\ref{app:constant-sum-proof} for the proof). All values in-between, $b \in (0\degree, 180\degree)$,  represent the spectrum of partially cooperative/competitive \textit{general-sum} games. This can be seen as an extension of the game from \citet{Crawford-Sobel-1982}, modified to cover the spectrum of cooperative/competitive scenarios.

\begin{figure}
    \begin{subfigure}{0.75\columnwidth}
    \begin{center}
    \includegraphics[width=\linewidth]{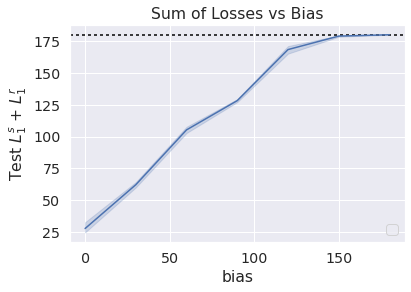}
    \end{center}
    \caption{}
    \label{fig:reinforce-deter}
    \end{subfigure}
    \begin{subfigure}{0.48\columnwidth}
    \begin{center}
    \includegraphics[width=\linewidth]{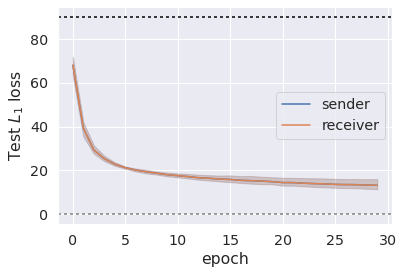}
    \end{center}
    \caption{$b = 0\degree$}
    \label{fig:cat-deter-0}
    \end{subfigure}
    \begin{subfigure}{0.48\columnwidth}
    \begin{center}
    \includegraphics[width=\linewidth]{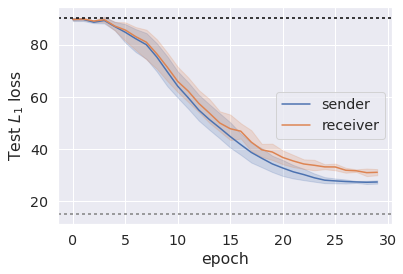}
    \end{center}
    \caption{$b = 30\degree$}
    \label{fig:cat-deter-3}
    \end{subfigure}
    \begin{subfigure}{0.48\columnwidth}
    \begin{center}
    \includegraphics[width=\linewidth]{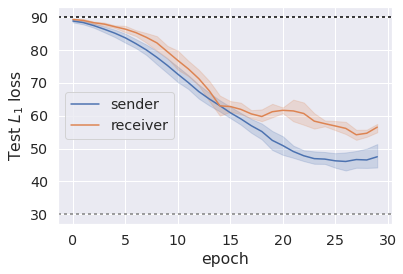}
    \end{center}
    \caption{$b = 60\degree$}
    \label{fig:cat-deter-6}
    \end{subfigure}
    \begin{subfigure}{0.48\columnwidth}
    \begin{center}
    \includegraphics[width=\linewidth]{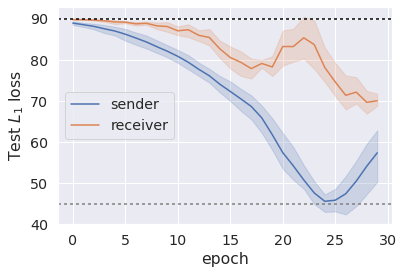}
    \end{center}
    \caption{$b = 90\degree$}
    \label{fig:cat-deter-9}
    \end{subfigure}
    \begin{subfigure}{0.48\columnwidth}
    \begin{center}
    \includegraphics[width=\linewidth]{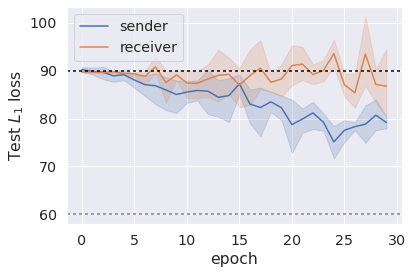}
    \end{center}
    \caption{$b = 120\degree$}
    \label{fig:cat-deter-12}
    \end{subfigure}
    \begin{subfigure}{0.48\columnwidth}
    \begin{center}
    \includegraphics[width=\linewidth]{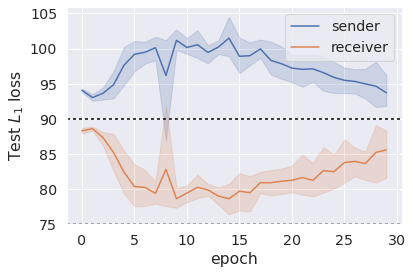}
    \end{center}
    \caption{$b = 150\degree$}
    \label{fig:cat-deter-15}
    \end{subfigure}
    \begin{subfigure}{0.48\columnwidth}
    \begin{center}
    \includegraphics[width=\linewidth]{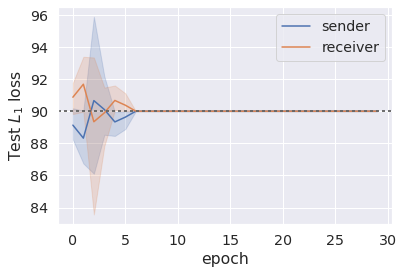}
    \end{center}
    \caption{$b = 180\degree$}
    \label{fig:cat-deter-18}
    \end{subfigure}
    \caption{(a)Lowest test loss ($L_1^r + L_1^s$) for the circular biased game found with a hyperparameter search for each $b \in [0, 30\degree, 60\degree, 90\degree, 120\degree, 150\degree, 180\degree]$. (b-h) Individual agent losses for the best run for each bias $b$. We plot the baseline loss under non-communication ($180\degree$) as a dark dotted line. Loss lower than non-communication baseline indicates effective communication.}
    \label{fig:cat-deter}
\end{figure}

\subsection{Training Details}
Both agents are implemented as MLPs with 2 hidden layers and ReLU \citep{Vinod2010ReLU} nonlinearities between layers. This was chosen as it is the smallest network that could reasonably solve the cooperative $b = 0\degree$ task (but we observed similar results with a 3 layer network as well). Optimization can be performed with either Gumbel-Softmax \citep{maddison2017-distribution99,jang2017-reparameterization28}\footnote{Since we train with independent agents that are competitive, we also thought Gumbel-Softmax would be inapproriate as it passes the receiver's gradient to the sender. This would mean the sender is backpropogating through the receiver's policy and is an unfair advantage.}, REINFORCE \citep{Williams1992SimpleSG} (also known as the score function estimator \citep{FuScoreFunc}) or a combination of the two where the receiver is updated with backpropogation and the sender with REINFORCE \citep{Schulman2015GradientEU} since the loss is differentiable with respect to the receiver. We choose the latter following previous work \citep{chaabouni2020-compositionality89, kharitonov2020-minimization60} and \citet{chaabouni2019-communication73} who show it converges more robustly. We train both agents independently and for REINFORCE we use an added entropy regularisation term, as well as a baseline for variance reduction\citep{Williams1992SimpleSG}.

We train for 30 epochs of 250 batches, with batch size 64, and set the circumference of our circle to 36 (so that a loss of $90\degree$ corresponds to a value $9$). Both agents are trained independently using Adam \citep{Kingma2014AdamAM}. To evaluate, we use a fixed test set of 100 equidistant points $\in [0\degree,180\degree]$ and take the expectation over the messages using the sender's distribution (as opposed to stochastically sampling a message during training). We do all hyperparameter searches with Or\'ion \citep{orion}, using random search with a fixed budget of 100 searches. We perform a hyperparameter search over both agents' learning rates, hidden layer sizes, the vocabulary size, and entropy regularisation (when used). We always report results averaged over 5 random seeds, plotting the mean over seeds as a line and a shaded area indicating 95\% confidence interval across seeds.
Since competitive MARL may not converge we take the average over the last 10 epochs to capture agents' final performance.\footnote{Our experiments generally never converge to one policy but have both agents dynamically maintain a final minimum loss nonetheless.} Hyperparameter search spaces and settings are described in supplementary~\ref{sup:hyperparameters}. Our model is implemented in PyTorch \citep{pytorch} and we use wandb \citep{wandb} for logging.

\section{Communication Is Proportional To Cooperation}

\subsection{Measuring Information Transfer}
To evaluate the effectiveness of the communication emerged, we can simply look to the sum of agents' $L_1$ losses. Under non-communication (or uninformative communication), we know that the receiver will just guess a point uniformally at random, so the expected loss for both agents is $\E_{\rx\sim U(0,360)} [ L_1^s(T_s,x)] = 90\degree $. Therefore, any error for either agent below $90\degree$ is evidence of information transfer \citep{Lowe2019OnTP}. Furthermore, since agents are not \textit{situated} there is no other action space which agents can use to communicate, all information transfer must be through the communication channel \citep{Mordatch-Abbeel-2017}. Therefore, the lower $L_1^r + L_1^s$, the more information is passed through the learned protocol.

\subsection{Emergent Communication for Different Levels of Competition}

Both \citet{Cao-et-al-2018,Jaques-et-al-2018} found that independent agents could not effectively learn to communicate in a competitive scenario and argue this is supported by \citet{Crawford-Sobel-1982}. Intuitively, though, the level of competition should affect whether communication is learned. To investigate, we use seven equidistant values of $b \in [0\degree, 30\degree, 60\degree, 90\degree, 120\degree, 150\degree, 180\degree]$ corresponding to different levels of competition. For each one, we do a hyperparameter search to find the lowest achievable sum of agents' errors, and empirically show whether communication can be learned under competition.  

We report our results in Figure~\ref{fig:cat-deter}. The overall results in Figure \ref{fig:reinforce-deter} show that the lowest achieved sum of errors is proportional to the bias between agents. Given that our agents perform better than the non-communicative baseline, we can conclude that agents can learn to communicate effectively without any special learning rules or losses contrary to current literature in machine learning. This matches the theoretical results of \citet{Crawford-Sobel-1982}; information transfer with communication is inversely proportional to the conflict of interest. We also confirm their theoretical result that effective communication is not possible past a certain bias --- in our case $b = 120\degree$, where information transfer becomes chaotic and negligible. We note that performance even in the fully cooperative $b = 0\degree$ case is not optimal because of the bottleneck of discrete communication and training instabilities of RL. We also show $b = 180\degree$, the constant-sum game ($L_1^s + L_1^r = 180\degree$) in Figure~\ref{fig:cat-deter-18}. Since the game is fully competitive, there is no incentive to communicate and all policies are uninformative after some initial instability. We show it here for completeness but, for brevity, we omit $b=180\degree$ in future experiments.

Overall, this empirical result extends the theoretical work of \citet{Crawford-Sobel-1982} to show that effective information transfer is not only theoretically possible, but also feasible to learn with the standard EC setup using RL. Furthermore, our results suggest that emergent communication agents can effectively learn communication proportional to the level of competition.

\section{Manipulation is not Communication}
\label{sec:manipulation}

\subsection{Information Transfer vs Communication}

While we have found evidence of information transfer, does that necessarily mean our agents have \textit{learned to communicate}? For example, if the sender were to be static and deterministic, the receiver could learn a perfect strategy to achieve 0 loss for itself. This would yield $L_1^r + L_1^s = b$ which is optimal information transfer and therefore evidence of communication according to \citet{Lowe2019OnTP}. But is this what we mean by \textit{communication}?

To further explore this question, we would like to show how a different model can seem to have better communication under the current metric but be inconsistent with our goals. For our different model, we change to using continuous, scalar messages similar to those used in a competitive setting in \citet{singh2018learning}. We make the sender's message a real-valued scalar and appropriately change its the distribution to be a 1D Gaussian, for which we now learn the mean and variance (see Algorithm \ref{alg:continuous-game} in supplementary). As with the previous experiment, we choose the minimal architecture that does well on the $b = 0\degree$ case, which is again an MLP with two hidden layers where the sender has two separate heads on top, one for predicting each of the mean and variance of the Gaussian.

\begin{figure}[t]
\begin{subfigure}{0.75\columnwidth}
\begin{center}
\includegraphics[width=\linewidth]{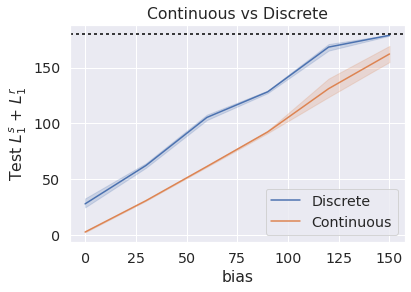}
\end{center}
\caption{}
\label{fig:discrete-v-continuous}
\end{subfigure}
\begin{subfigure}{0.48\columnwidth}
\begin{center}
\includegraphics[width=\linewidth]{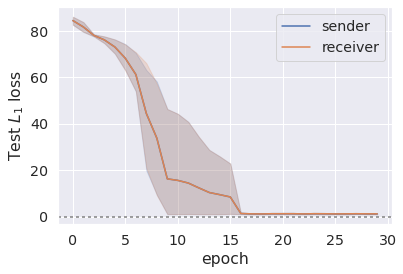}
\end{center}
\caption{$b = 0\degree$}
\label{fig:gauss-deter-bias0}
\end{subfigure}
\begin{subfigure}{0.48\columnwidth}
\begin{center}
\includegraphics[width=\linewidth]{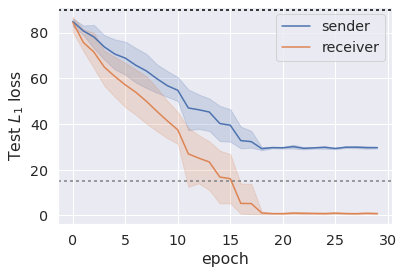}
\end{center}
\caption{$b = 30\degree$}
\label{fig:gauss-deter-bias3}
\end{subfigure}
\begin{subfigure}{0.48\columnwidth}
\begin{center}
\includegraphics[width=\linewidth]{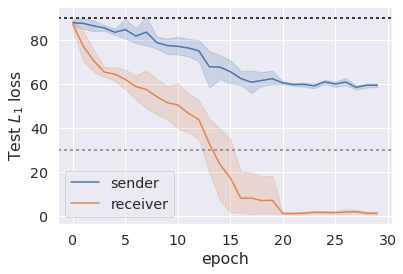}
\end{center}
\caption{$b = 60\degree$}
\label{fig:gauss-deter-bias6}
\end{subfigure}
\begin{subfigure}{0.48\columnwidth}
\begin{center}
\includegraphics[width=\linewidth]{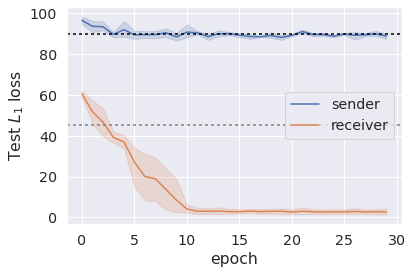}
\end{center}
\caption{$b = 90\degree$}
\label{fig:gauss-deter-bias9}
\end{subfigure}
\begin{subfigure}{0.48\columnwidth}
\begin{center}
\includegraphics[width=\linewidth]{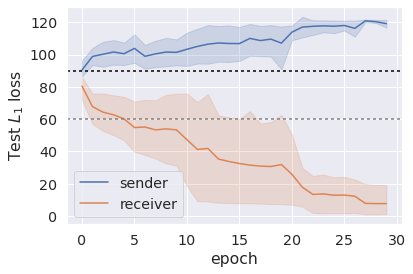}
\end{center}
\caption{$b = 120\degree$}
\label{fig:gauss-deter-bias12}
\end{subfigure}
\begin{subfigure}{0.48\columnwidth}
\begin{center}
\includegraphics[width=\linewidth]{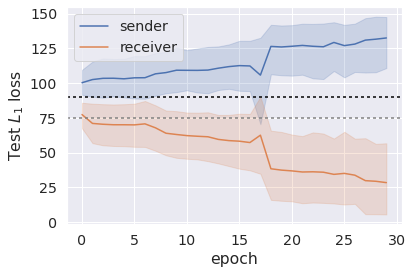}
\end{center}
\caption{$b = 150\degree$}
\label{fig:gauss-deter-bias15}
\end{subfigure}
\caption{(a) Total test loss over biases $b \in [30\degree, 60\degree, 90\degree, 120\degree]$ in the circular biased game for in a setup with continuous messages compared to discrete messages. (b-g) Individual agent reward for the best runs found for continuous messages for each bias. We plot the loss under non-communication ($90\degree$, dark dotted line) and loss for an even split of the bias ($b/2$, light dotted line) as baselines. Loss higher than the non-communication baseline indicates manipulation.}
\label{fig:gauss-deter}
\end{figure}

We re-run our experiments with continuous messages and report our results in Figure~\ref{fig:gauss-deter}. The overall results in Figure~\ref{fig:discrete-v-continuous} suggest that the protocols learned with continuous messages are all highly informative and better than those with discrete messages. However, looking at the best runs for highly competitive cases $b=120\degree,150\degree$ in Figures~\ref{fig:gauss-deter-bias12}, \ref{fig:gauss-deter-bias15}, we find that the receiver dominates the sender, performing near-optimally while the sender's loss is usually close to the bias $b$. We can deduce that the receiver manages to guess its own target correctly. This is not advantageous to the sender, as it performs worse than if it were not communicating at all (dark dotted line). This suggests that the sender is not choosing to transfer information about the target but rather the receiver is reading it from the sender's message despite the sender's best efforts. Though it is information transfer, this does not resemble communication and requires us to consider literature in signalling \citep{Skyrms-2010-Signals} to delineate the differences.

\subsection{Communication vs Manipulation}
\label{sub:our-metric}
The literature in signalling refers to information transfer as either \textit{communication} \citep{Barrett-Skyrms-2017} or \textit{manipulation} of receivers by senders \citep{Dawkins-Krebs-1978} or vice-versa \citep{Hinde-1981}. The latter manifests as one agent dominating the other and is modelled as {\it cue-reading} or {\it sensory manipulation}. Communication, as opposed to manipulation, requires \textit{both} agents to receive a net benefit \citep{Krebs-Dawkins-1984}, and this implies some degree of cooperation \citep{Lewis-1969}. For fully-cooperative emergent communication, previous metrics of joint reward \citep{Lowe2019OnTP}, or causal influence \citep{Lowe2019OnTP, Jaques-et-al-2018}, are sufficient to find information transfer and can therefore be used as effective metrics. But for competitive scenarios, neither of these can distinguish between manipulation and cooperation \citep{Skyrms-Barrett-2018} and using them as metrics (e.g. \citep{Jaques-et-al-2018}) risks finding manipulation and confusing it for communication.

\subsection{Measuring Competitive Communication}
\begin{figure}[t]
\begin{subfigure}{0.48\columnwidth}
\begin{center}
\includegraphics[width=\linewidth]{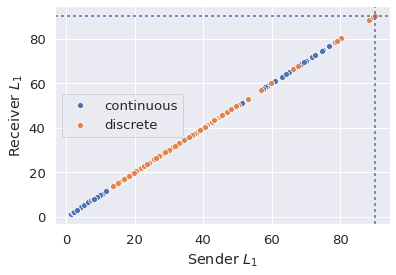}
\end{center}
\caption{$b = 0\degree$}
\label{fig:cont-discrete-bias0}
\end{subfigure}
\begin{subfigure}{0.48\columnwidth}
\begin{center}
\includegraphics[width=\linewidth]{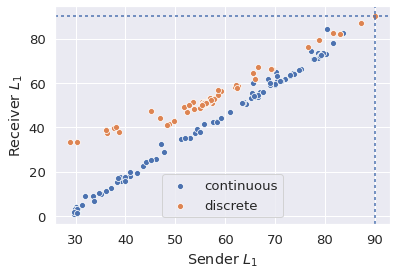}
\end{center}
\caption{$b = 30\degree$}
\label{fig:cont-discrete-bias3}
\end{subfigure}
\begin{subfigure}{0.48\columnwidth}
\begin{center}
\includegraphics[width=\linewidth]{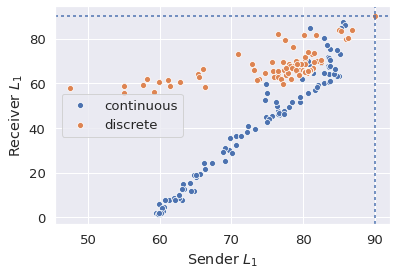}
\end{center}
\caption{$b = 60\degree$}
\label{fig:cont-discrete-bias6}
\end{subfigure}
\begin{subfigure}{0.48\columnwidth}
\begin{center}
\includegraphics[width=\linewidth]{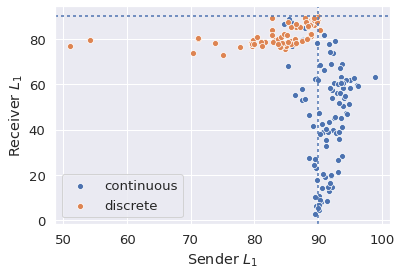}
\end{center}
\caption{$b = 90\degree$}
\label{fig:cont-discrete-bias9}
\end{subfigure}
\begin{subfigure}{0.48\columnwidth}
\begin{center}
\includegraphics[width=\linewidth]{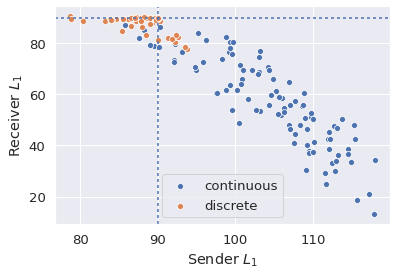}
\end{center}
\caption{$b = 120\degree$}
\label{fig:cont-discrete-bias12}
\end{subfigure}
\begin{subfigure}{0.48\columnwidth}
\begin{center}
\includegraphics[width=\linewidth]{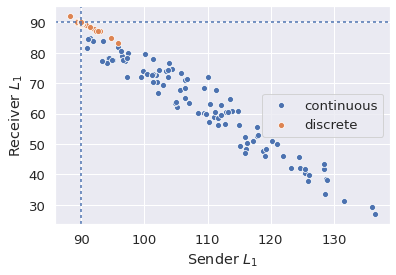}
\end{center}
\caption{$b = 150\degree$}
\label{fig:cont-discrete-bias15}
\end{subfigure}
\caption{Receiver vs Sender loss for all 100 hyperparameter runs for bias (a) $0\degree$ (b) $30\degree$ (c) $60\degree$ (d) $90\degree$ (e) $120\degree$ (f) $150\degree$ in the continuous and discrete message setups. We plot the baseline loss under non-communication ($L_1 = 90\degree$) as a dotted line. Loss higher than the non-communication baseline indicates manipulation. We report the loss averaged over 5 seeds.}
\label{fig:hyperparam-discrete-cont}
\end{figure}

To accurately compare the inductive biases our models of discrete and continuous messages, we plot the results of all 100 hyperparameter searches instead of just choosing the best one. For each bias $b$ we compare every hyperparameter configuration of our discrete and continuous message models using the $L_1$ error for each agent. We report our results in Figure~\ref{fig:hyperparam-discrete-cont}.

Since our focus is on the emergence of \textit{communication}, we are looking for settings where both agents benefit: they perform better than either their fully-exploited losses ($L_1^s < b$ and $L_1^r < b$) or the loss under non-communication ($L_1^s < 90\degree$ and $L_1^r < 90\degree$). We see throughout that our discrete model is relatively fair. In contrast, the receiver with continuous messages dominates the sender. Points follow a general slope of $L_1^s = L_1^r + b$ regardless of the bias $b$. In competitive scenarios $b = 120\degree,150\degree$ in Figure \ref{fig:cont-discrete-bias12},\ref{fig:cont-discrete-bias15}, the sender with continuous messages has very few runs that performs better than non-communication. All other runs clearly show manipulation by the receiver. This corroborates our intuition that this model with continuous messages is predisposed to finding manipulation, not communication.\footnote{This model is only used to demonstrate how manipulation can be mistaken for communication. A different setup with continuous messages can learn strong non-manipulative communication (see  supplementary \ref{sup:cont-fair}) and a discrete message system can learn manipulation (see certain runs in Figure~\ref{fig:cont-discrete-bias12},\ref{fig:cont-discrete-bias15})} On the other hand, our model with discrete messages has both agents perform similarly well and therefore biases towards communication. Though the receiver usually does better, we find that the difference between agents' losses is not overly large and indicative of widespread manipulation, like we see with continuous messages.

This demonstrates the importance of qualitatively evaluating our models' biases but eventually, we need to choose the best hyperparameters and this requires a quantitative metric of competitive communication. Previous approaches\citep{Jaques-et-al-2018} used the simple sum of agents' losses but we propose the sum of squared losses $L_2 = (L_1^s)^2+(L_1^r)^2$ as a better heuristic metric. We can view our partially competitive scenario as having a \textit{common-interest loss} ($180\degree - b$), in which both agents are fully cooperative, and a \textit{conflict-of-interest loss} ($b$), in which both agents are fully competitive. The sum of $L_1$ losses optimises only for the common interest, whereas $L_2$ prefers a more fair division of the conflict-of-interest loss in addition to optimising common interest (see supplementary~\ref{app:fairness-proof} for proof).

An important note is that this $L_2$ metric need only be used for choosing the best hyperparameters. Within our game, we can still use the $L_1$ (or another metric) as the loss function. As well, we stress that this metric is still just a heuristic and the best run found may still exhibit manipulation (as is the case in our setup continuous message), and it is important to verify that both agents cooperatively benefit from communication. Furthermore, given the extensive use of hyperparameter optimization in RL \citep{henderson2018deep}, we believe our qualitative hyperparameter plot is an effective way to find models with good biases towards communication. Finally, we note that we can extend other cooperative communication metrics \citep{Lowe2019OnTP,Jaques-et-al-2018} to competitive scenarios by adding the condition that both agents benefit from communication.

\section{Communication Requires Both Agents to Possibly Benefit}

\subsection{The Negotiation Game}
If both agents must benefit for communication to emerge, what happens if one agent is already at its optimal error? \citet{Cao-et-al-2018} proposed a negotiation game to study EC between agents $A$ and $B$ that are either fully-cooperative and collectively optimise total reward (`pro-social', reward-sharing) or competitive and independently optimise their own reward (`selfish'). We study their `selfish' setup\footnote{`Prosocial' agents are fully-cooperative so they, by default, can learn to communicate optimally since communication is proportional to cooperation} which they find fails to learn to communicate. They suggest the reason is either the game is too competitive \citep{Crawford-Sobel-1982} or not iterated. However, investigating more closely, the issue may be due to one agent being better off not communicating.

We describe the game briefly here; for a full description please see the original paper. A pool of 3 types of items with quantity up to 5, $i \in \{0,\ldots,5\}^3$ is initialised for a game between two agents. Each agent $j$ is assigned a private utility $u_j \in \{0,\ldots,10\}^3$ over each item and they negotiate with each other to split the items and try to gain a favourable split for themselves. At every turn $t$, an agent creates a proposal for how to split up the objects $p_j^t \in \{0,\ldots,5\}^3$. If the other party accepts and the proposal is valid, agents are rewarded with $R_j = p_j^t \cdot u_j$ and $R_{1-j} = (i - p_j^t) \cdot u_j$. If the proposal is invalid, both agents get 0 reward. If an agent refuses the proposal, it gives a counter-proposal and the game continues until either a proposal is accepted or the game goes past a limit $N$ and both agents get reward 0. Agent $A$ always goes first. \citet{lazaridou2020-communication82} describe it as a modified, multi-turn version of the ultimatum game \citep{Guth-et-al-1982}.

\citet{Cao-et-al-2018} experimented with four possible communication configurations. In the first case (`non-communication'), agents do not communicate at all, so neither agent can see the other's proposals. In the second case (`proposal'), agents can see each other's proposals. In the third case (`linguistic'), agents can send each other a cheap talk message consisting of a sequence of tokens from a vocabulary. This corresponds to the emergent communication case, and allows agents to learn to communicate either their proposal, utilities, or other information. The final configuration (`both') corresponds to agents having access to both the linguistic and the proposal channels of communication.

\citet{Cao-et-al-2018} find that for the `non-communication' setting, agent $B$ soundly dominates, taking nearly all the items. For the `linguistic' setting, agents manage to split the item-pool relatively fairly but do not converge and exhibit chaotic dynamics. For the `proposal' and `both' settings, agents manage to converge to policies that split the item-pool perfectly evenly.

\subsection{Reproducing and Investigating}

\begin{figure}[t!]
    \centering
    \subcaptionbox{Non-communication \label{fig:self-none}}{
        \includegraphics[width=0.48\columnwidth]{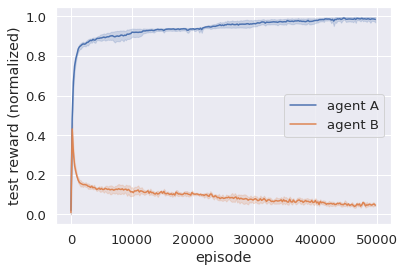}}
    \hfill
    \subcaptionbox{Linguistic \label{fig:self-ling}}{
        \includegraphics[width=0.48\columnwidth]{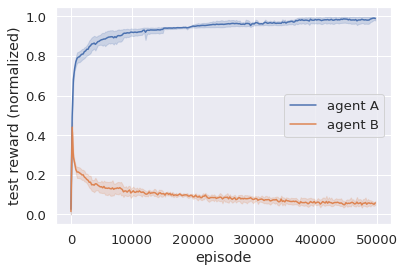}}
    \hfill
    \subcaptionbox{Proposal \label{fig:self-prop}}{
        \includegraphics[width=0.48\columnwidth]{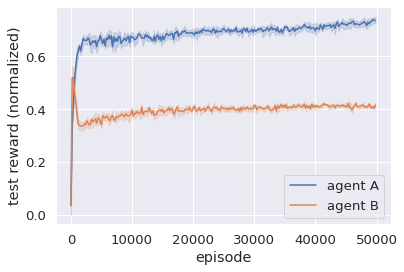}}
    \hfill
    \subcaptionbox{Both \label{fig:self-both}}{
        \includegraphics[width=0.48\columnwidth]{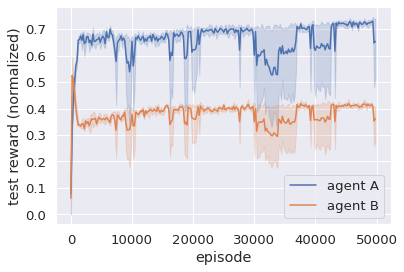}}
    \hfill

    \caption{Test reward per agent in our reproduction of the `selfish' negotiation game \citep{Cao-et-al-2018} for a setup (a) without communication (b) with only the linguistic channel (c) with only the proposal channel (d) with both channels. We plot the mean and a $95\%$ confidence interval over the seeds. Following the original paper, we report the reward divided by the total possible reward an agent could achieve.
    }
    \label{fig:ecn}
\end{figure}

We reproduce the experiments and report our results in Figure~\ref{fig:ecn}. We largely manage to reproduce the results for the `proposal' and `both' settings, with the small difference that the reward is not perfectly split and agent $A$ is now winning slightly. For the `non-communication' setting our results are nearly identical except again, agent $A$ is dominating instead of $B$. We believe this is reasonable as it matches results on the aforementioned ultimatum game \citep{Guth-et-al-1982} that gives the first-mover ($A$) an advantage. Stochastically varying the number of available turns $N$, as \citet{Cao-et-al-2018} propose, did not alleviate the first-mover advantage in our experiments.

\begin{figure}
    \centering
    \subcaptionbox{Masked-Linguistic Reward \label{fig:self-ling-masking-reward}}{
        \includegraphics[width=0.48\columnwidth]{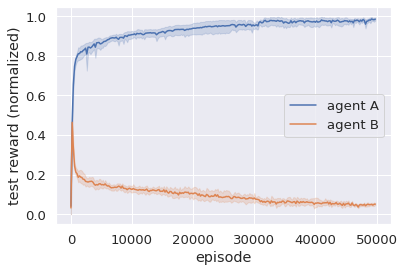}}
    \hfill
    \subcaptionbox{$\%$ message unmasked \label{fig:self-ling-masking-mask}}{
        \includegraphics[width=0.48\columnwidth]{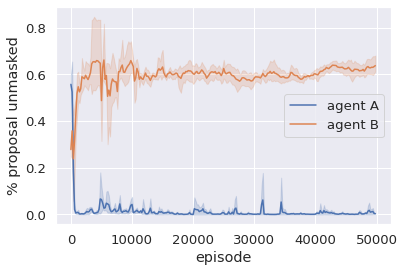}}
    \caption{The negotiation game using the masked-linguistic channel (a) reward per agent (b) percent of messages each agent unmasked}
    \label{fig:masked-ling}
\end{figure}

However, we find noticeably different results for the `linguistic' settings despite our best efforts.\footnote{Code for the original paper is not publicly accessible, so we contacted and worked with the original authors. We spent several weeks investigating after already having a working codebase but could not reproduce the results. All our code will be open-sourced and made available online.}
Although \citet{Cao-et-al-2018} find that the game with a linguistic channel has both agents split the item pool equally, we find that the game is essentially unchanged from the `non-communication' case. Our results in Figure~\ref{fig:self-ling} are nearly identical to those in Figure~\ref{fig:self-none}.

\subsection{Agents Learn Not To Communicate}

To discover the reason for the discrepancy in results, we probe the behaviour of our agents. One possibility is that the game is too difficult to learn and there is an issue with optimization. To simplify, we use the linguistic policy to generating a binary mask  $m_j \in \{0,1\}^3$ (instead of a sequence of tokens). We then set the agent's linguistic message to the masked version of their proposal $m_j \cdot u_j$. Since proposals are grounded in the game (they determine the reward split), and learning a binary mask is much easier than a sequence with tokens from a vocabulary, we expect optimization issues to be greatly reduced. As well, this `masked-linguistic' messaging system allows agents to explicitly hide or reveal their proposal to their opponent. In this way, we can probe whether the behaviour of an agent is to hide information or to share it.

 We report our results in Figure~\ref{fig:masked-ling} which shows that the reward (Figure~\ref{fig:self-ling-masking-reward}) with this channel is unchanged compared to the original `linguistic' setup. This implies that agents are not having difficulty learning to communicate as it is incredibly easy to learn a mask $[1,1,1]$ that would communicate their proposal precisely. Instead, looking at the percent of messages each agent unmasked (Figure~\ref{fig:self-ling-masking-mask}), it is clear agent $B$ seems to be trying to communicate but agent $A$ is specifically masking everything. Agent $A$ explcitly chooses not to communicate their proposal.

 This is rational, because if agent $A$ dominates in the case of non-communication, it is in their best interest to stay in that setup. By masking their proposal, they reduce this setting to the `non-communication' setup as opposed to the `proposal' setup where their reward is less. \citet{Cao-et-al-2018} suggest that agents {\it fail} to learn effective negotiation. Instead, we suggest that agent $A$ specifically {\it chooses} to suppress any possible communication, preventing negotiation. This also suggests that our results in reproducing the negotiation game are more plausible as, in the original work, adding the cheap talk `linguistic' channel causes the game to go from domination to a fair split.

From this we can infer that if an agent that has nothing to gain from communication, it should not be expected to learn to communicate. In particular, agents that dominate in a non-communicative setting (`none') but do worse when information is exchanged (`proposal') will explicitly learn not to communicate.

\section{\textit{Competitive} Emergent Communication Through Negotiation}
We have demonstrated three different results for competitive emergent communication but what started this inquiry was the specific claim that independent agents could not learn to effectively communicate in the negotiation game \citep{Cao-et-al-2018}. We now examine the negotiation game from the perspective of our three takeaways. The game is not zero-sum as long as agents have different utilities for each item. Even though agents' interest overlap, different utilities guarantee that the game is general-sum so communication should be non-zero. An agent that manipulates instead of communicating would take all the items and this would be easy to spot. The only remaining issue is that agent A dominates in the non-communication case, making learning to communicate infeasible. By fixing the non-communication case, we should be able to learn to communicate in this game.

To alleviate the non-communication domination, we can change the conditions of termination to even the playing field. In the original work, agents explicitly output a termination signal that agrees to their opponent's last proposal but agent B too easily learns to agree to whatever proposal it is given. We change the game to ignore the termination output and instead end the game when an agent outputs a proposal that is the same as their opponent's last proposal, $p_j^t == p_{1-j}^{t-1}$. This way, learning to terminate the game becomes more difficult and agent B is harder to take advantage of. This should also make it more difficult to dominate in the non-communication case where you cannot communicate your proposal.

\begin{figure}
    \centering
    \begin{subfigure}{0.48\columnwidth}
    \begin{center}
    \includegraphics[width=\linewidth]{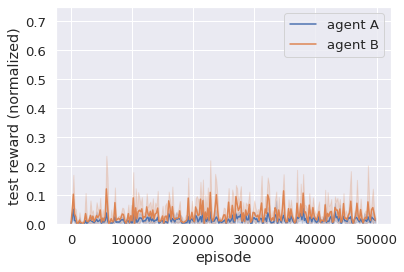}
    \end{center}
    \caption{Non-communication}
    \label{fig:self-none-propterm}
    \end{subfigure}
    \begin{subfigure}{0.48\columnwidth}
    \begin{center}
    \includegraphics[width=\linewidth]{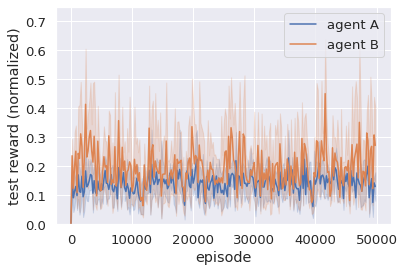}
    \end{center}
    \caption{Linguistic}
    \label{fig:self-ling-propterm}
    \end{subfigure}
    \begin{subfigure}{0.48\columnwidth}
    \begin{center}
    \includegraphics[width=\linewidth]{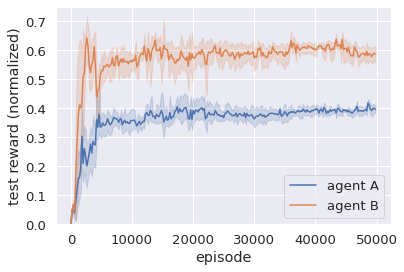}
    \end{center}
    \caption{Proposal}
    \label{fig:self-prop-propterm}
    \end{subfigure}
    \begin{subfigure}{0.48\columnwidth}
    \begin{center}
    \includegraphics[width=\linewidth]{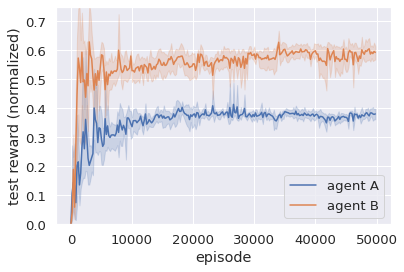}
    \end{center}
    \caption{Both}
    \label{fig:self-both-propterm}
    \end{subfigure}
    \caption{Reward in the negotiation game with termination through proposal agreement for setups (a) without communication (b) only linguistic channel (c) only proposal channel (d) both linguistic and proposal channels}
    \label{fig:self-propterm}
\end{figure}

We report the results of using proposal termination in Figure~\ref{fig:self-propterm}. For `proposal' and `both', we see similar reward patterns to the original game except now agent $B$ is slightly better than agent $A$. More importantly, though, is that the non-communication setting is \textit{equally} badly for both agents and \textit{both} are incentivised to communicate and improve their reward. With `linguistic' channel agents successfully learn to communicate where they previous had not. Though the dynamics are slightly chaotic, the average reward for each agent is clearly above the non-communication baseline demonstrating effective information transfer. We also see both agents benefit from the communication so we can be sure this is not manipulation. The total reward utilization is low but this can be explained by the highly competitive nature of the negotiation game. Since agents always have overlapping interests, depending on the utilities each agent has for each item, the game may be much closer to fully-competitive than fully-cooperative. Overall, with a small modification we have shown that even in the negotiation game, competitive agents can learn to communicate.

\section{Conclusion}

Importantly, we have demonstrated the first clear, positive result of emergent communication under competition in modern machine learning using standard RL. We have expanded on conditions for competitively learning to communicate and presented three results. (1) We provide evidence that information transfer is proportional to cooperation, applying the theoretical results of \citet{Crawford-Sobel-1982} to modern emergent communication. (2) We highlight a subtle distinction between \textit{information transfer}, \textit{communication}, and \textit{manipulation}, and demonstrated qualitative and quantitative approaches to differentiating the last two. We argue that you can extend metrics of cooperative emergent communication \citep{Lowe2019OnTP} to competitive cases with the caveat that both agents must benefit from communication. This is especially useful for reflecting on the metric used to choose the best hyperparameters in general-sum games \citep{Jaques-et-al-2018}. (3) We demonstrate that \textit{both} agents must be able to benefit from communication for it to be feasible. To do so, we reproduced one of the first investigations into competitive emergent communication, the negotiation game \citep{Cao-et-al-2018}, and clarified their hypothesis for why communication did not emerge. Finally, we have combined our three takeaways to argue there is only a single issue preventing emergent communication in the negotiation game. We have gone back to this first, modern negative result of competitive emergent communication and with a small change, we have successfully learned to negotiate between competitive agents.

Modern emergent communication can be seen through two overarching lines of research: understanding the properties of emergent languages and using emergent communication to improve AI systems \citep{lazaridou2020-communication82}. Research in both of these goals has been fruitful. However, it is usually limited to the case of fully-cooperative, reward-sharing agents which may be restricting to the fields and enforce a (false) perspective that communication must be fully cooperative. In contrast, other fields have already been using it as a tool for cooperation and coordination in partially-competitive environments \citep{Crawford-Sobel-1982, Farrell-Rabin-1996}.

We believe that competitive emergent communication can be useful to both avenues of research. In understanding emergent languages, competition can be another dimension to investigate which protocols emerge---e.g., how competition may regularise a community of speakers. Another line of research could investigate how more complex protocols change with the level of competition. For improving AI systems, competitive emergent communication offer a more realistic perspective on human communication. For agents to understand that how human communication can lie or hide the truth, they must be at least partially competitive. And, for communication protocols to be resilient to lying, they may also need to be trained with competitive agents.

Our hope is that this paper invigorates research into competitive emergent communication and highlights some key takeaways for future research to progress this new and exciting field.

\FloatBarrier



\begin{acks}
MN would like to thank Adrien Ali Ta\"iga,  David Krueger, Ryan Lowe, and previous conference reviewers for comments and feedback on earlier versions of this work. TL would like to thank Schwartz Reisman Institute at the University of Toronto for partially funding this research.
\end{acks}



\bibliographystyle{ACM-Reference-Format}
\bibliography{main}

\newpage
\FloatBarrier
\appendix

\section{Hyperparameters}
\label{sup:hyperparameters}
For the experiments in Section 4,5 we run a random search over the following hyperparameter search space. For compactness we write $\{S,R\}$ below, but note that the sender and receiver's hyperparameters are distinct and randomly drawn separately. See \texttt{configs/} directory of the code repository for more information including the best hyperparameters found for each setup.

\begin{align*}
    \text{vocabulary size } |V| &\sim [64,128,256] \\
    \text{learning rate } \alpha_{S,R} &\sim \log \text{Uniform}(10^{-4}, 10^{-2}) \\
    \text{hidden layer size } h_{S,R} &\sim \text{Uniform}(16, 64) \\
    \text{entropy regularization } \lambda_{S,R} &\sim \log \text{Uniform}(10^{-4}, 1)
\end{align*}

For Section 6,7, we used the same hyperparameters as in the original paper
\begin{align*}
    \text{vocabulary size } |V| &= 11 \\
    \text{max utterance length} |m| &= 6 \\
    \text{learning rate } \alpha_{S,R} &= \text{ADAM defaults} \\
    \text{hidden size } h_{S,R} &= 100 \\
    \text{termination entropy coeff } \lambda_{S,R}^{term} &= 0.05 \\
    \text{proposal entropy reg coeff } \lambda_{S,R}^{prop} &= 0.05 \\
    \text{utterance entropy reg coeff } \lambda_{S,R}^{utt} &= 0.001 \\
    \text{number of rounds } N &\sim Poisson(7) \text{ in range} [4,10]
\end{align*}

\section{Proofs}
\subsection{Proof of Fully Cooperative/Fully Competitive Game}
\label{app:constant-sum-proof}
For $b = 0$, $T_s = T_r$ so trivially $L_s = L_r$ and the game is fully cooperative.

\begin{figure}
\begin{center}
    \includegraphics[scale=0.3]{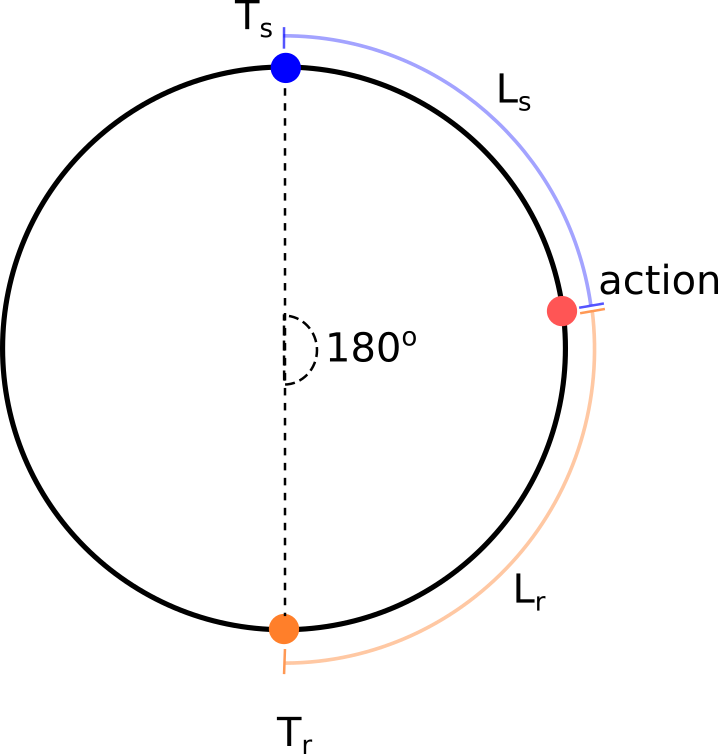}
\end{center}
\caption{The game with maximal bias $180\degree$ showing the sum of $L_1$ losses $L_r + L_s = 180\degree$}
\label{fig:sender-receiver-180}
\end{figure}

For $b = 180\degree, T_r = T_s + 180 \mod 360\degree$  we provide a visual demonstration in Figure \ref{fig:sender-receiver-180} that the sum is always $L_s + L_r = 180\degree$ and therefore the game is constant-sum and fully competitive. We can also think of this as moving the actions distance $d$ towards one agent's target means moving it distance $d$ away from the other agent's target.

$0 \leq  T_r, T_s, a \leq 360\degree$. Assume without loss of generality $T_s < T_r$ so $T_r = T_s + 180\degree$ and $T_s \leq 180\degree \leq T_r \leq 360 \degree$
\begin{align*}
L_s + L_r &= L_1(T_s, a) + L_1(T_r, a) \\
&= \min(|T_s - a|, 360\degree - |T_s - a|) \\
&+ \min(|T_r - a|, 360\degree - |T_r - a|) \\
\end{align*}
\begin{align*}
\intertext{\textbf{case 1}: $|T_s - a| < 360\degree - |T_s - a|$}
\min(|T_s - a|, 360\degree - |T_s - a|) &= |T_s - a| \\
|T_s - a| &< 180\degree \\
\intertext{\textbf{subcase a}: $T_s >= a$}
|T_s - a| &= T_s -a \\
\because T_r &= T_s + 180\degree \\
T_r &\geq a + 180\degree \\
\therefore \min(|T_r - a|,  360\degree - |T_r - a|) &= 360 - (T_r - a) \\
L_s + L_r &=  T_s - a + 360\degree - (T_r - a) \\
&= T_s - a + 360\degree - T_s - 180\degree + a \\
&= 180
\intertext{\textbf{subcase b}: $T_s < a$}
|T_s - a| &= a - T_s \\
a - T_s &< 180\degree \\
a &< T_s + 180\degree \\
a &< T_r \\
\therefore |T_r - a| &= T_r - a \\
\because T_r& = T_s + 180\degree \\
T_r &< a + 180 \degree \\
T_r -a &< 180\degree \\
2(T_r - a) &< 360 \degree \\
T_r - a &< 360\degree - (T_r - a) \\
\therefore \min(|T_r - a|,  360\degree - |T_r - a|) &= T_r - a \\
\therefore L_s + L_r &=  (a - T_s) + (T_r - a) \\
&= T_r - T_s \\
&= 180 \degree
\end{align*}

We can extend the proof by symmetry (on the circle) for  $|T_s - a| \geq 360\degree - |T_s - a|$, so the sum of losses $L_r + L_s$ always equals $180\degree$ so the game is constant-sum and therefore fully competitive.

\subsection{Proof of $L_2$ Fairness}
\label{app:fairness-proof}
Assume without loss of generality $T_s < T_r$, we are minimizing the sum of $L_2$ losses

\begin{align*}
\min_a L_s + L_r &= \min_a (T_s - a)^2 + (T_r - a)^2 \\
&= \min_a (T_s - a)^2 + (T_s + b - a)^2 \\
\intertext{let $T_s = x$}
&= \min_a (x - a)^2 + (x + b - a)^2 \\
&= \min_a x^2 - 2ax + a^2 + x^2 + 2bx \\
&+ b^2 - 2ax - 2ab + a^2 \\
&= \min_a 2(x - 2ax + a^2 + bx - ab + b^2/4) \\
&+ b^2/2 \\
&= \min_a 2(x - a + b/2)^2 + b^2/2 \\
a &= x + b/2
\end{align*}

Sum of $L_2$ losses is minimised when the action is $T_s + b/2$ or halfway between both agents' targets.


\begin{algorithm}
\caption{Circular Biased Sender-Receiver Game with Discrete Messages}
\label{alg:game}
\begin{algorithmic}
\Procedure{Training Batch}{$b$}
\State $T_s \sim  \text{Uniform}(0,360)$
\State $T_r \gets T_s + b$
\State $m \sim \text{Categorical}(S(T_s))$
\State $a \gets R(m)$
\State $L_s \gets L_{1}(T_s,a) = \min(|T_s - a|, 360 -|T_s - a|)$
\State $L_r \gets L_{1}(T_r,a) = \min(|T_r - a|, 360 -|T_r - a|)$
\State $R$ is updated with SGD
\State $S$ is updated with REINFORCE
\EndProcedure
\end{algorithmic}
\end{algorithm}

\begin{algorithm}
\caption{Circular Biased Sender-Receiver Game with Continuous Messages}
\label{alg:continuous-game}
\begin{algorithmic}
\Procedure{Training Batch}{$b$}
\State $T_s \sim  \text{Uniform}(0,360)$
\State $T_r \gets T_s + b$
\State $\mu, \sigma \gets S(T_s)$
\State $m \sim \text{Gaussian}(\mu, \sigma)$
\State $a \gets R(m)$
\State $L_s \gets L_{1}(T_s,a) = \min(|T_s - a|, 360 -|T_s - a|)$
\State $L_r \gets L_{1}(T_r,a) = \min(|T_r - a|, 360 -|T_r - a|)$
\State $R$ is updated with SGD
\State $S$ is updated with REINFORCE
\EndProcedure
\end{algorithmic}
\end{algorithm}

\newpage

\begin{figure}[t]
\begin{subfigure}{0.78\columnwidth}
\begin{center}
\includegraphics[width=\linewidth]{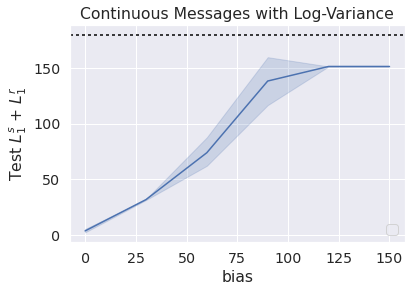}
\end{center}
\caption{}
\label{fig:vargauss-all}
\end{subfigure}
\begin{subfigure}{0.48\columnwidth}
\begin{center}
\includegraphics[width=\linewidth]{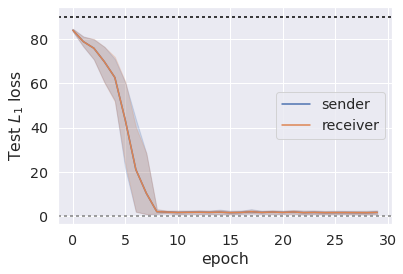}
\end{center}
\caption{$b = 0\degree$}
\label{fig:vargauss-deter-bias0}
\end{subfigure}
\begin{subfigure}{0.48\columnwidth}
\begin{center}
\includegraphics[width=\linewidth]{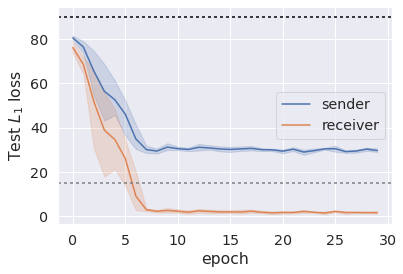}
\end{center}
\caption{$b = 30\degree$}
\label{fig:vargauss-deter-bias3}
\end{subfigure}
\begin{subfigure}{0.48\columnwidth}
\begin{center}
\includegraphics[width=\linewidth]{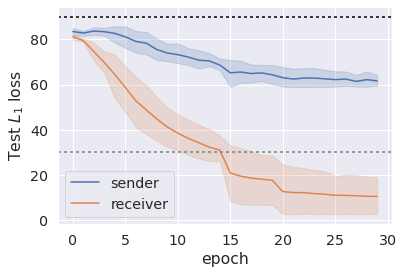}
\end{center}
\caption{$b = 60\degree$}
\label{fig:vargauss-deter-bias6}
\end{subfigure}
\begin{subfigure}{0.48\columnwidth}
\begin{center}
\includegraphics[width=\linewidth]{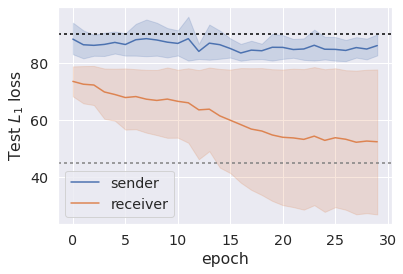}
\end{center}
\caption{$b = 90\degree$}
\label{fig:vargauss-deter-bias9}
\end{subfigure}
\begin{subfigure}{0.48\columnwidth}
\begin{center}
\includegraphics[width=\linewidth]{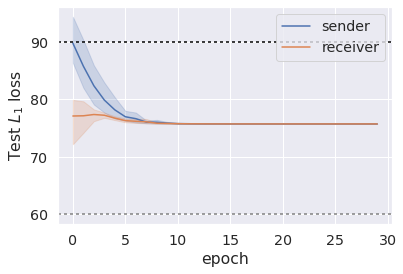}
\end{center}
\caption{$b = 120\degree$}
\label{fig:vargauss-deter-bias12}
\end{subfigure}
\begin{subfigure}{0.48\columnwidth}
\begin{center}
\includegraphics[width=\linewidth]{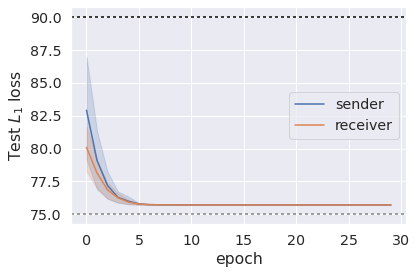}
\end{center}
\caption{$b = 150\degree$}
\label{fig:vargauss-deter-bias15}
\end{subfigure}
\caption{(a) Total test loss over biases $b \in [30\degree, 60\degree, 90\degree, 120\degree]$ in the circular biased game for in a setup with continuous messages compared to discrete messages. (b-g) Individual agent reward for the best runs found for continuous messages for each bias. We plot the loss under non-communication ($90\degree$, dark dotted line) and loss for an even split of the bias ($b/2$, light dotted line) as baselines. Loss higher than the non-communication baseline indicates manipulation.}
\label{fig:vargauss-deter}
\end{figure}

\section{Non-Manipulative Continuous}
\label{sup:cont-fair}

Section~\ref{sec:manipulation} looks at a model with continuous messages that exhibits manipulation instead of communication. This model is only to demonstrate how manipulation can be mistaken for communication but is not about arguing the superiority of discrete messages. For completeness, we show here how a continuous message model can learn fair and effective communication.

In the original continuous message model we learn the mean and variance of a Gaussian from which we sample our messages. For our fair model, we instead learn the mean and \textit{log-variance} of our Gaussian. We plot the results of the best hyperparameters found for each bias in Figure~\ref{fig:vargauss-deter}. We see a clear difference between the `log-variance' results and original `variance' results. In the original runs, the receiver had complete control of the game and minimised its loss at the expense of the sender. Here, we see a relative even split, even for the highly competitive $b=120\degree$ case (Figure~\ref{fig:gauss-deter-bias12}).

\begin{figure}
    \centering
     \begin{subfigure}[b]{0.46\columnwidth}
         \begin{center}
         \includegraphics[width=\linewidth]{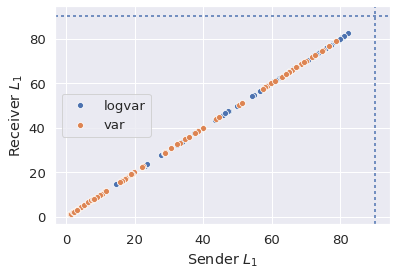}
         \end{center}
         \caption{$b = 0\degree$}
     \end{subfigure}
     \begin{subfigure}[b]{0.46\columnwidth}
         \begin{center}
         \includegraphics[width=\linewidth]{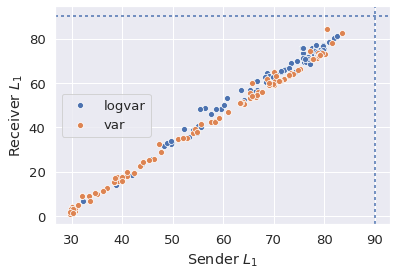}
         \end{center}
         \caption{$b = 30\degree$}
     \end{subfigure}
     \begin{subfigure}[b]{0.46\columnwidth}
         \centering
         \includegraphics[width=\linewidth]{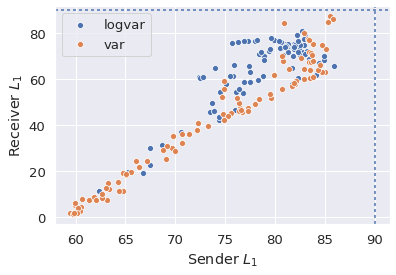}
         \caption{$b = 60\degree$}
         \label{fig:var_v_logvar_bias6}
     \end{subfigure}
      \begin{subfigure}[b]{0.46\columnwidth}
         \centering
         \includegraphics[width=\linewidth]{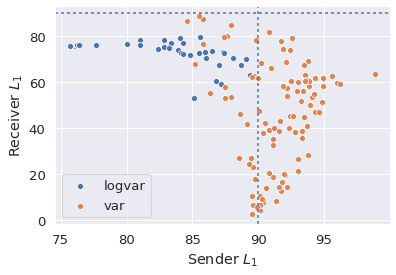}
         \caption{$b = 90\degree$}
         \label{fig:var_v_logvar_bias9}
     \end{subfigure}
     \begin{subfigure}[b]{0.46\columnwidth}
         \centering
         \includegraphics[width=\linewidth]{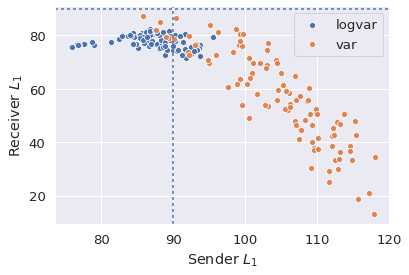}
         \caption{$b = 120\degree$}
        \label{fig:var_v_logvar_bias12}
     \end{subfigure}
     \begin{subfigure}[b]{0.46\columnwidth}
         \centering
         \includegraphics[width=\linewidth]{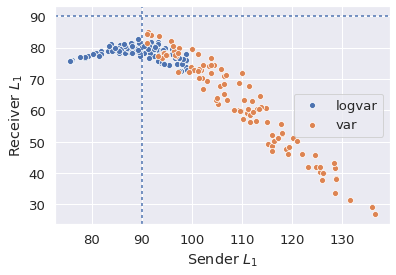}
         \caption{$b = 150\degree$}
         \label{fig:var_v_logvar_bias15}
     \end{subfigure}
    \caption{Receiver vs Sender loss for all 100 hyperparameter runs for bias (a) $0\degree$ (b) $30\degree$ (c) $60\degree$ (d) $90\degree$ (e) $120\degree$ (f) $150\degree$ in the continuous variance (`var', orange) and log-variance (`logvar', blue) setups. We plot the baseline loss under non-communication ($L_1 = 90\degree$) as a dotted line. Loss higher than the non-communication baseline indicates manipulation.}
    \label{fig:var_v_logvar}
\end{figure}

Investigating further, we plot all the hyperparameter searches for our regular variance (`var') Gaussian and the new log-variance (`logvar') in Figure~\ref{fig:var_v_logvar}. Compared to the original variance setup, the log-variance setup has much fewer instances of manipulation and agent losses are more fairly distributed. In highly competitive cases, $b = 120\degree,150\degree$ (Figure~\ref{fig:var_v_logvar_bias12},\ref{fig:var_v_logvar_bias15}) we not only see much fairer losses but also much better performance overall

This implies that learning the variance of a Gaussian with REINFORCE is sub-optimal compared to learning the log-variance and mediocre sender performance may be explained by optimization difficulties.


\end{document}